\documentclass[runningheads,a4paper]{llncs}

\usepackage{amssymb}
\usepackage{graphicx}

\usepackage{amsmath,amssymb}

\newcommand{\htransf}{H}

\usepackage{url}

\begin{document}

\mainmatter  % start of an individual contribution

\title{Separable time-causal and time-recursive spatio-temporal receptive fields} 

\titlerunning{Separable time-causal and time-recursive spatio-temporal receptive fields}

\authorrunning{Tony Lindeberg}

\author{Tony Lindeberg\thanks{Support from Swedish Research
    Council contracts 2010-4766 and 2014-4083 is gratefully acknowledged}}

\institute{Department of Computational Biology\\
School of Computer Science and Communication\\ KTH Royal
Institute of Technology, Stockholm, Sweden}

\maketitle
<\begin{abstract}
We present an improved model and theory for time-causal and
time-recursive spatio-temporal receptive fields,
obtained by a combination of Gaussian receptive fields over the
spatial domain and first-order integrators or equivalently truncated
exponential filters coupled in cascade over the temporal domain.
Compared to previous spatio-temporal scale-space formulations in terms
of non-enhancement of local extrema or scale invariance, these
receptive fields are based on different scale-space axiomatics over
time by ensuring non-creation of new local extrema or
zero-crossings with increasing temporal scale.
Specifically, extensions are presented about parameterizing the intermediate
temporal scale levels, analysing the resulting temporal dynamics and transferring
the theory to a discrete implementation in terms of recursive filters
over time.
\end{abstract}

\section{Introduction}

Spatio-temporal receptive fields constitute an essential concept in
biological vision 
(Hubel and Wiesel \cite{HubWie05-book}; 
 DeAngelis et al.\ \cite{DeAngOhzFre95-TINS,deAngAnz04-VisNeuroSci})
and for expressing computer vision methods on video data
(Adelson and Bergen \cite{AdeBer85-JOSA}; 
 Zelnik-Manor and Irani \cite{ZelIra01-CVPR}; 
 Laptev and Lindeberg \cite{LapLin04-ECCVWS};
 Jhuang et al.\ \cite{JhuSerWolPog07-ICCV};
 Shabani et al.\ \cite{ShaClaZel12-BMVC}).
For off-line processing of pre-recorded video, non-causal Gaussian or
Gabor-based spatio-temporal receptive
fields may in some cases be sufficient.
When operating on video data in a real-time setting or when modelling
biological vision computationally, one does however need to take into explicit
account the fact that the future cannot be accessed and that the 
underlying spatio-temporal receptive fields must be
{\em time-causal\/}.
For computational efficiency and for keeping down memory requirements,
it is also desirable that the computations should be {\em time-recursive\/},
so that it is sufficient to keep a limited memory of the past that can
be recursively updated over time.

The subject of this article is to present an improved scale-space
model for spatio-temporal receptive fields based on time-causal
temporal scale-space kernels in terms of first-order integrators
coupled in cascade, which can also be transferred to a discrete
implementation in terms of recursive filters.
The model builds on previous work by 
(Fleet and Langley \cite{FleLan95-PAMI};
 Lindeberg and Fagerstr{\"o}m \cite{LF96-ECCV};
 Lindeberg \cite{Lin10-JMIV,Lin13-BICY}) which is here complemented by a better
 design for the degrees of freedom in the choice of time constants for intermediate
temporal scale levels, an analysis of the resulting temporal response dynamics and
details for discrete implementation in a spatio-temporal visual front-end.
%
% Conceptually, our approach is also related to the
% time-causal scale-time model by Koenderink \cite{Koe88-BC} which is here
% complemented by a truly time-recursive formulation of time-causal receptive
% fields more suitable for real-time operations over a compact temporal
% buffer of what has occurred in the past including a theoretically
% well-founded and computationally efficient method for discrete implementation.

\section{Spatio-temporal receptive fields}
\label{sec-spat-temp-RF}

The theoretical structure that we start from is a general result 
from axiomatic derivations of a spatio-temporal scale-space
based on assumptions of non-enhancement of local extrema
and the existence of a continuous temporal scale parameter,
which states that the spatio-temporal receptive fields should be based on
spatio-temporal smoothing kernels of the form
(see overviews in Lindeberg \cite{Lin10-JMIV,Lin13-BICY}):
\begin{equation}
       T(x_1, x_2, t;\; s, \tau, v, \Sigma)  
       = g(x_1 - v_1 t, x_2 - v_2 t;\; s, \Sigma) \, h(t;\; \tau)
      \label{eq-spat-temp-RF-model}
\end{equation}
  where
  \begin{itemize}
  \item
     $x = (x_1, x_2)^T$ denotes the image coordinates,
  \item
     $t$ denotes time,
  \item
    $s$ denotes the spatial scale,
\item
    $\tau$ denotes the temporal scale,
  \item
   $v = (v_1, v_2)^T$ denotes a local image velocity,
  \item
    $\Sigma$ denotes a spatial covariance matrix determining the
    spatial shape of an affine Gaussian kernel
   $g(x;\; s, \Sigma)  = \frac{1}{2 \pi s \sqrt{\det\Sigma}} e^{-x^T \Sigma^{-1} x/2s}$,
  \item
     $g(x_1 - v_1 t, x_2 - v_2 t;\; s, \Sigma)$ denotes a spatial affine Gaussian kernel
     that moves with image velocity $v = (v_1, v_2)$ in space-time and
\item
   $h(t;\; \tau)$ is a temporal smoothing kernel over time.
\end{itemize}
For simplicity, we shall here restrict the family of affine Gaussian
kernels over the spatial domain to rotationally symmetric Gaussians of
different size $s$, by setting the covariance matrix $\Sigma$ to a unit
matrix.
We shall mainly restrict ourselves to space-time separable
receptive fields by setting the image velocity $v$ to zero.

A conceptual difference that we shall pursue is by relaxing the
requirement of a continuous temporal scale parameter in the above
axiomatic derivations by a discrete temporal scale parameter.
We shall also replace the previous axiom about non-creation of new
image structures with increasing scale in terms of non-enhancement of
local extrema (which requires a continuous scale parameter) by the
requirement that the temporal smoothing process, when seen as an operation
along a one-dimensional temporal axis only, must not increase the number of
local extrema or zero-crossings in the signal. Then, another family of
time-causal scale-space kernels becomes permissible and uniquely determined, in terms of
first-order integrators or truncated exponential filters
coupled in cascade. 

The main topics of this paper are to handle the remaining degrees of freedom
resulting from this construction about: (i)~choosing and parameterizing
the distribution of temporal scale levels, (ii)~analysing the resulting temporal
dynamics and (iii)~describing how this model can be transferred to a
discrete implementation while retaining discrete scale-space
properties.

\section{Time-causal temporal scale-space}
\label{sec-time-caus-scale-spaces}

When constructing a system for real-time processing of sensory data,
a fundamental constraint on the temporal smoothing
kernels is that they have to be {\em time-causal\/}. 
The ad hoc solution of using a truncated
symmetric filter of finite temporal extent in combination with a
temporal delay is not appropriate in a time-critical context.
Because of computational and memory efficiency, the computations
should furthermore be based on a compact temporal buffer that contains
sufficient information for representing sensory information at multiple
temporal scales and computing features therefrom.
Corresponding requirements are necessary in computational
modelling of biological perception.

\paragraph{Time-causal scale-space kernels for pure temporal domain.}

Given the requirement on temporal scale-space kernels by
non-creation of local extrema over a pure temporal domain,
{\em truncated exponential kernels\/}
  \begin{equation}
    h_{exp}(t;\; \mu_k) 
    = \left\{
        \begin{array}{ll}
          \frac{1}{\mu_k} e^{-t/\mu_k} & t \geq 0 \\
          0         & t < 0
        \end{array}
      \right.
  \end{equation}
can be shown to constitute the only
class of time-causal scale-space kernels over a continuous temporal domain in
this sense
(Lindeberg \cite{Lin90-PAMI}; Lindeberg and Fagerstr{\"o}m \cite{LF96-ECCV}).
The Laplace transform of such a kernel is given by
\begin{equation}
    H_{exp}(q;\; \mu_k) 
    = \int_{t = - \infty}^{\infty} h_{exp}(t;\; \mu_k) \, e^{-qt} \, dt
    = \frac{1}{1 + \mu_k q}
  \end{equation}
  and coupling $K$ such kernels in cascade leads to a composed filter
  \begin{equation}
    \label{eq-comp-trunc-exp-cascade}
    h_{composed}(t;\; \mu) 
    = *_{k=1}^{K} h_{exp}(t;\; \mu_k)
  \end{equation}
  having a Laplace transform of the form
  \begin{align}
    \begin{split}
       \label{eq-expr-comp-kern-trunc-exp-filters}
       H_{composed}(q;\; \mu) 
        = \int_{t = - \infty}^{\infty} (*_{k=1}^{K} h_{exp}(t;\; \mu_k)) \, e^{-qt} \, dt
        =  \prod_{k=1}^{K} \frac{1}{1 + \mu_k q}.
    \end{split}
  \end{align}
  The composed filter has temporal mean and variance
  \begin{equation}
     \label{eq-mean-var-trunc-exp-filters}
     m_K = \sum_{k=1}^{K} \mu_k \quad\quad \tau_K = \sum_{k=1}^{K} \mu_k^2.
  \end{equation}
In terms of physical models, repeated convolution with such kernels
corresponds to coupling a series
of {\em first-order integrators\/} with time constants $\mu_k$ in cascade
\begin{equation}
   \partial_t L(t;\; \tau_k) 
   = \frac{1}{\mu_k} \left( L(t;\; \tau_{k-1}) - L(t;\; \tau_k) \right)
\end{equation}
with $L(t;\; 0) = f(t)$.
These temporal smoothing kernels satisfy scale-space properties in
  the sense that the number of local extrema or the number of zero-crossings
  in the temporal signal are guaranteed to not increase with
  the temporal scale.
  In this respect, these kernels have a desirable and well-founded smoothing
  property that can be used for defining multi-scale observations over time.
  A constraint on this type of temporal scale-space representation,
  however, is that the {\em scale levels are required to be discrete\/}
  and that the scale-space representation does hence not admit
  a continuous scale parameter.
 Computationally, however, the scale-space representation based
  on truncated exponential kernels can be highly efficient and admits
  for direct implementation in terms of hardware (or wetware) that emulates
  first-order integration over time, and where the temporal scale
  levels together also serve as a sufficient time-recursive memory of the past.

\begin{figure}[hbt]
  \begin{center}
    \begin{tabular}{ccc}
      $h(t;\; \mu, K=7)$ & $h_{t}(t;\; \mu, K=7)$ & $h_{tt}(t;\; \mu, K=7)$ \\
      \includegraphics[width=0.22\textwidth]{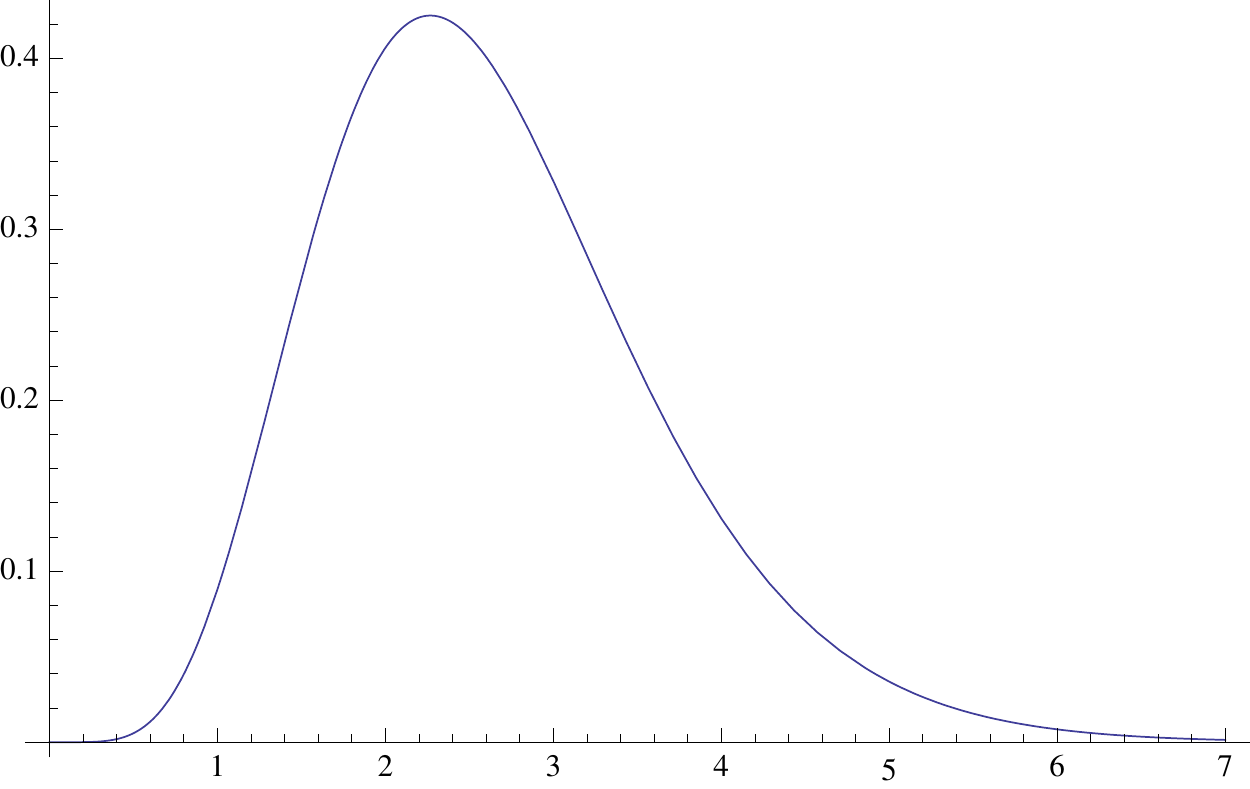} &
      \includegraphics[width=0.22\textwidth]{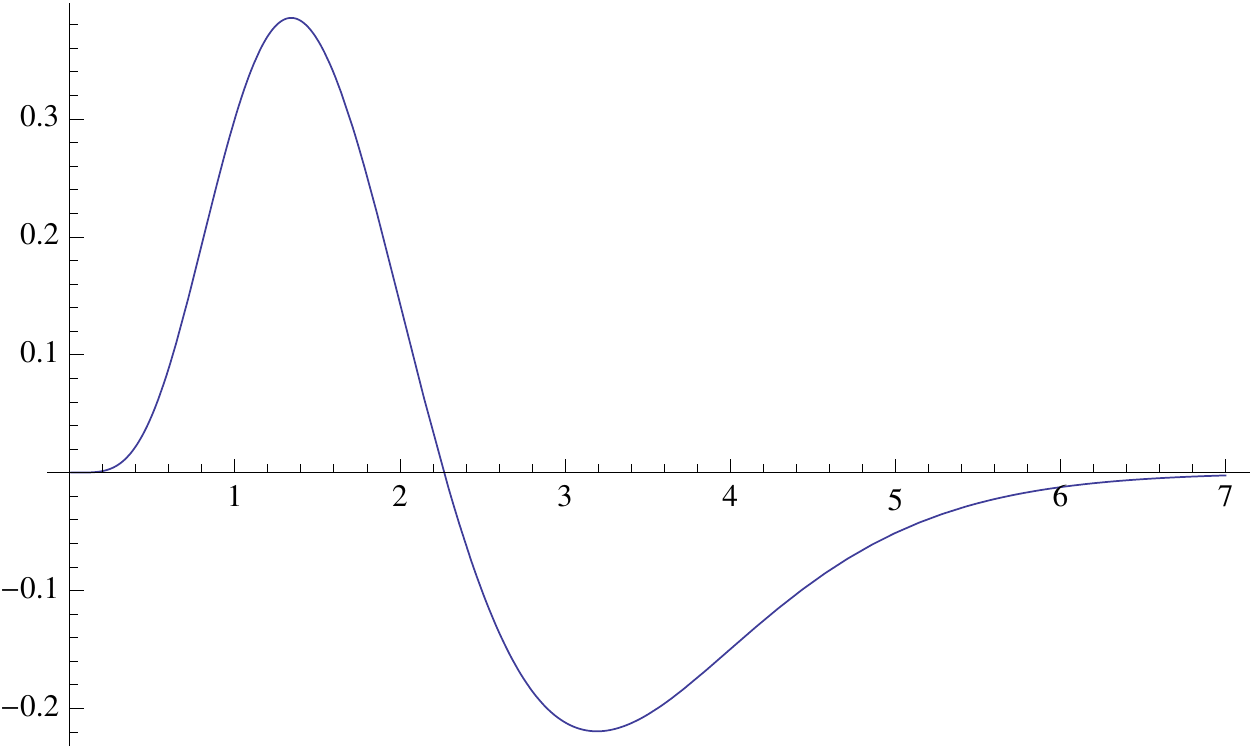} &
      \includegraphics[width=0.22\textwidth]{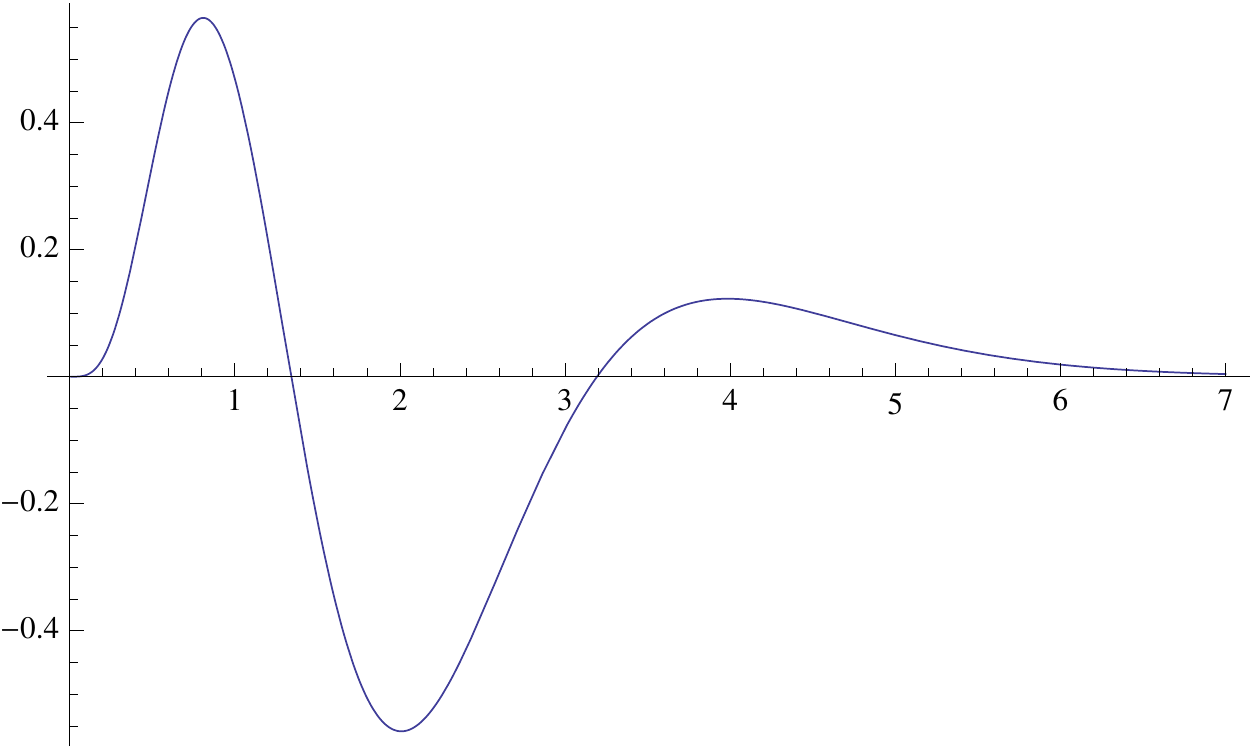} \\
  \\
      {\small $h(t;\; K=7, c = \sqrt{2})$} 
      & {\small $h_{t}(t;\; K=7, c = \sqrt{2}))$} 
      & {\small $h_{tt}(t;\; K=7, c = \sqrt{2}))$} \\
      \includegraphics[width=0.22\textwidth]{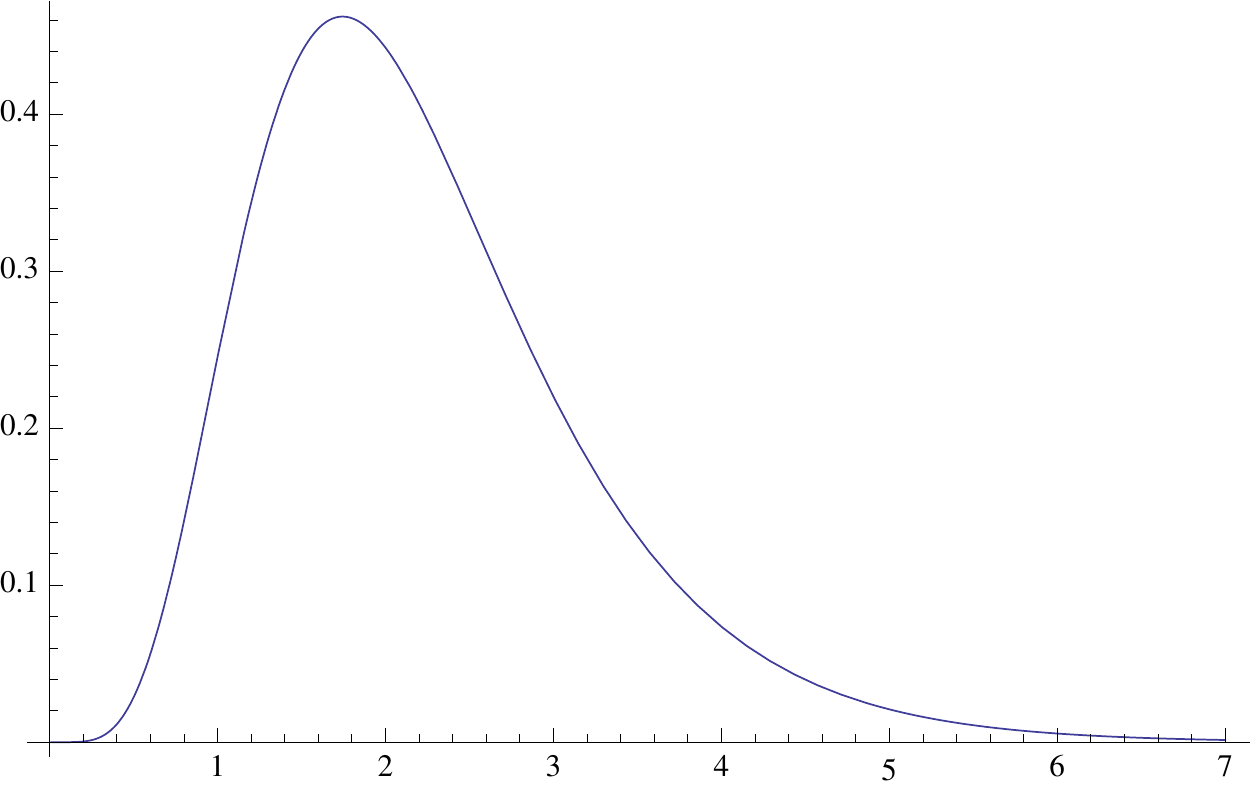} &
      \includegraphics[width=0.22\textwidth]{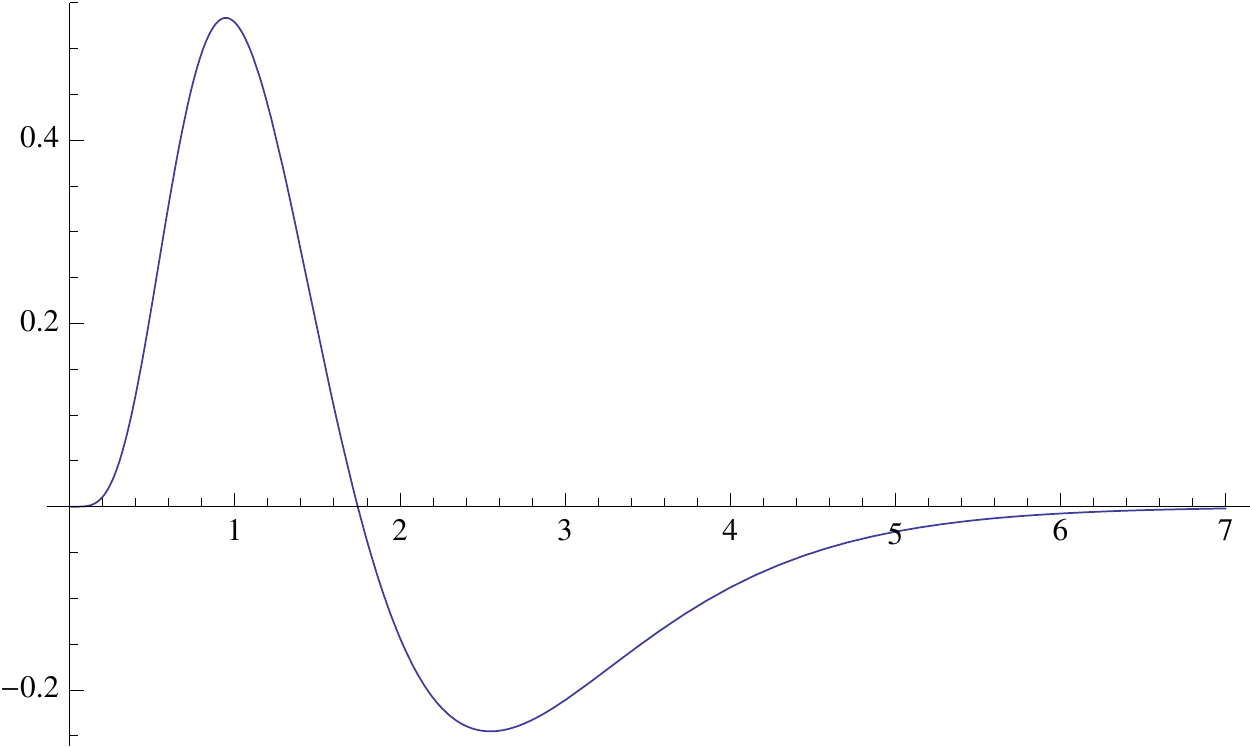} &
      \includegraphics[width=0.22\textwidth]{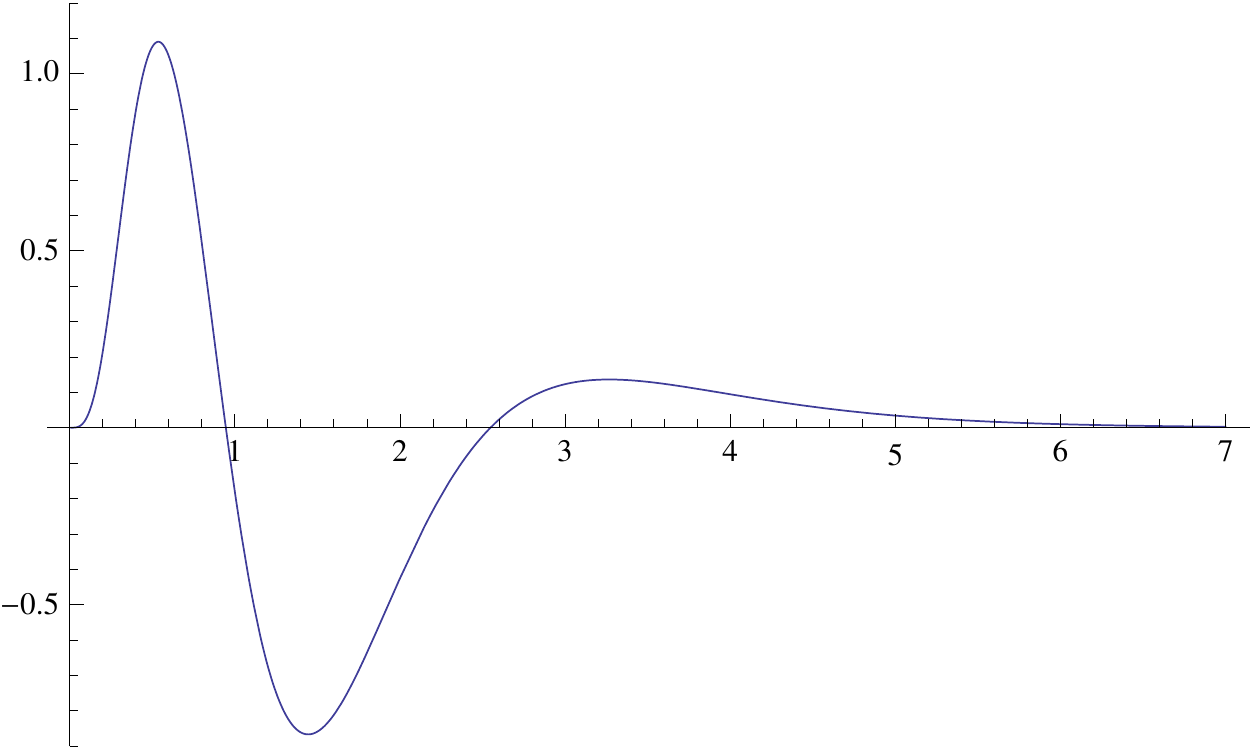} \\
\\
      {\small $h(t;\; K=7, c = 2^{3/4})$} 
      & {\small $h_{t}(t;\; K=7, c = 2^{3/4}))$} 
      & {\small $h_{tt}(t;\; K=7, c = 2^{3/4}))$} \\
      \includegraphics[width=0.22\textwidth]{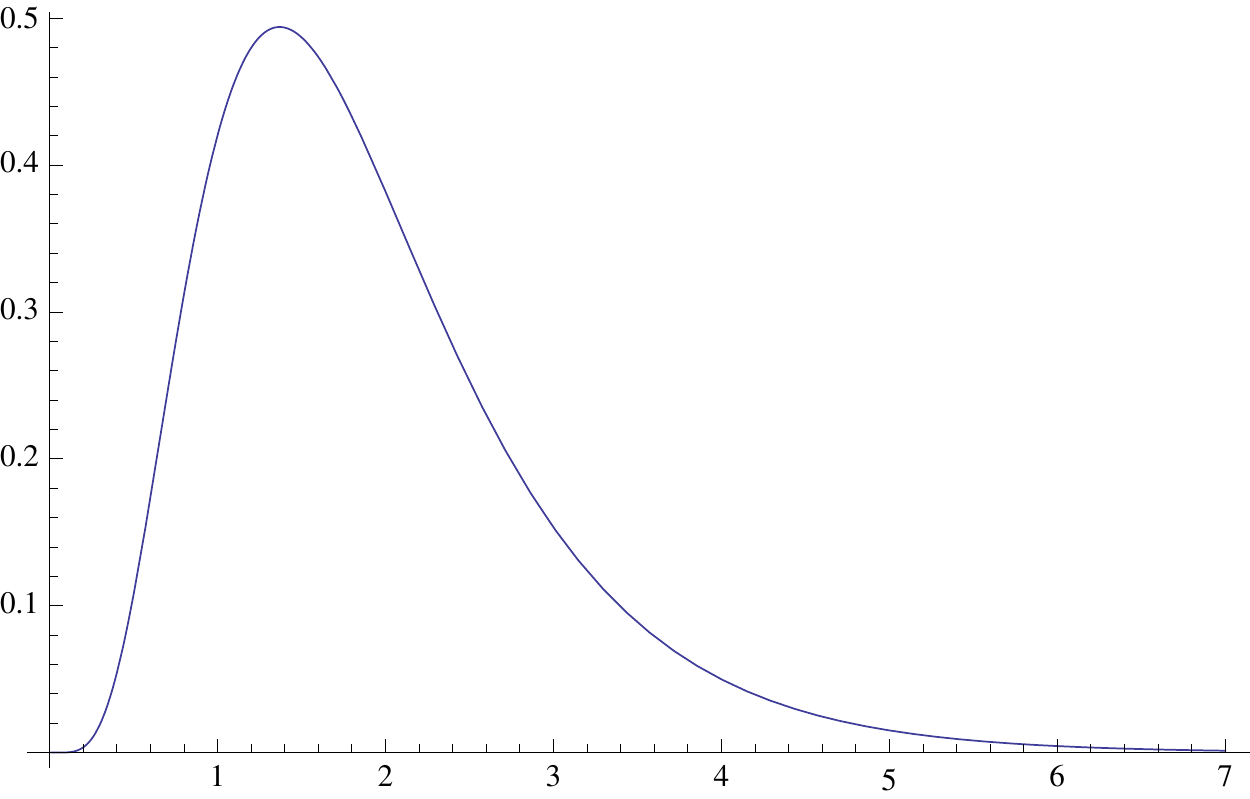} &
      \includegraphics[width=0.22\textwidth]{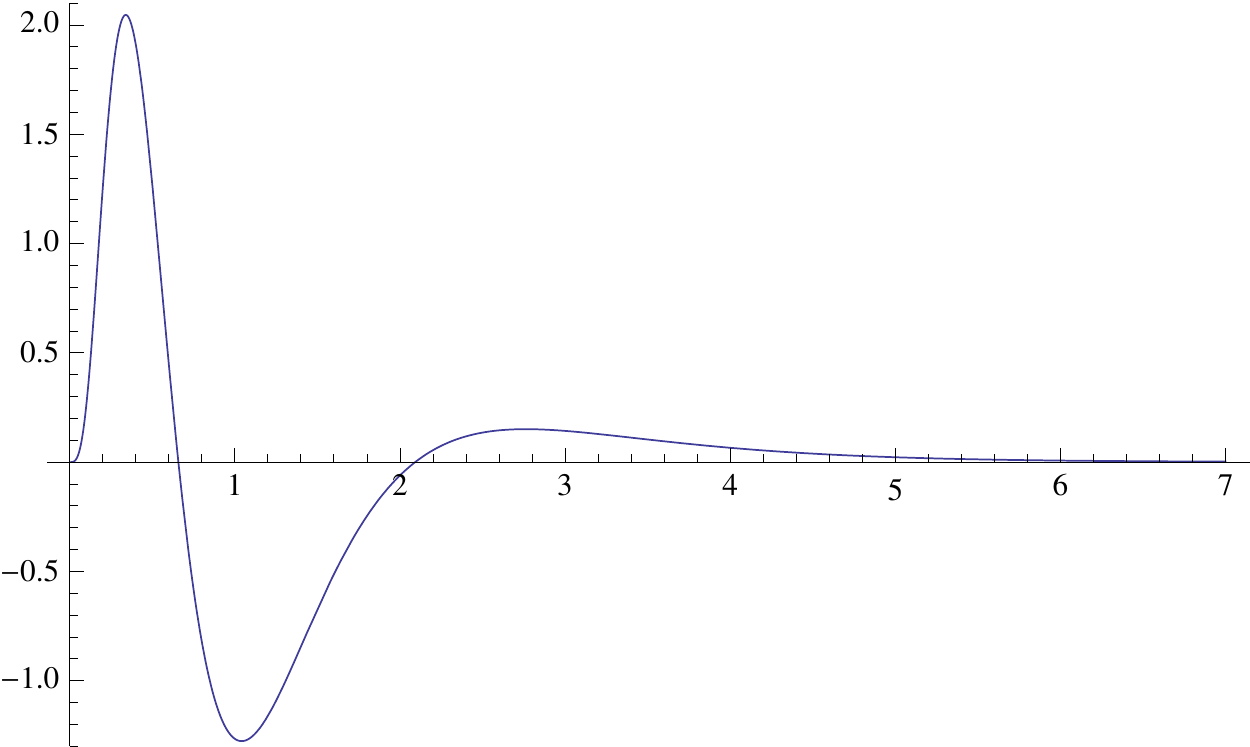} &
      \includegraphics[width=0.22\textwidth]{truncexp7logkernel2r0p75-dtt-eps-converted-to.pdf} \\
    \end{tabular} 
  \end{center}
   \caption{Equivalent kernels with temporal variance $\tau = 1$ corresponding to the composition of
           $K$ {\em truncated exponential kernels\/} in cascade and
           their first- and second-order derivatives.
           (top row) Equal time constants $\mu$.
           (middle row) Logarithmic distribution of the scale levels
           for $c = \sqrt{2}$.
          (bottom row) Logarithmic distribution 
           for $c = 2^{3/4}$.}
  \label{fig-trunc-exp-kernels-1D}
\end{figure}

\begin{figure}[hbt]
  \begin{center}
     \begin{tabular}{cc}
        \includegraphics[width=0.50\textwidth]{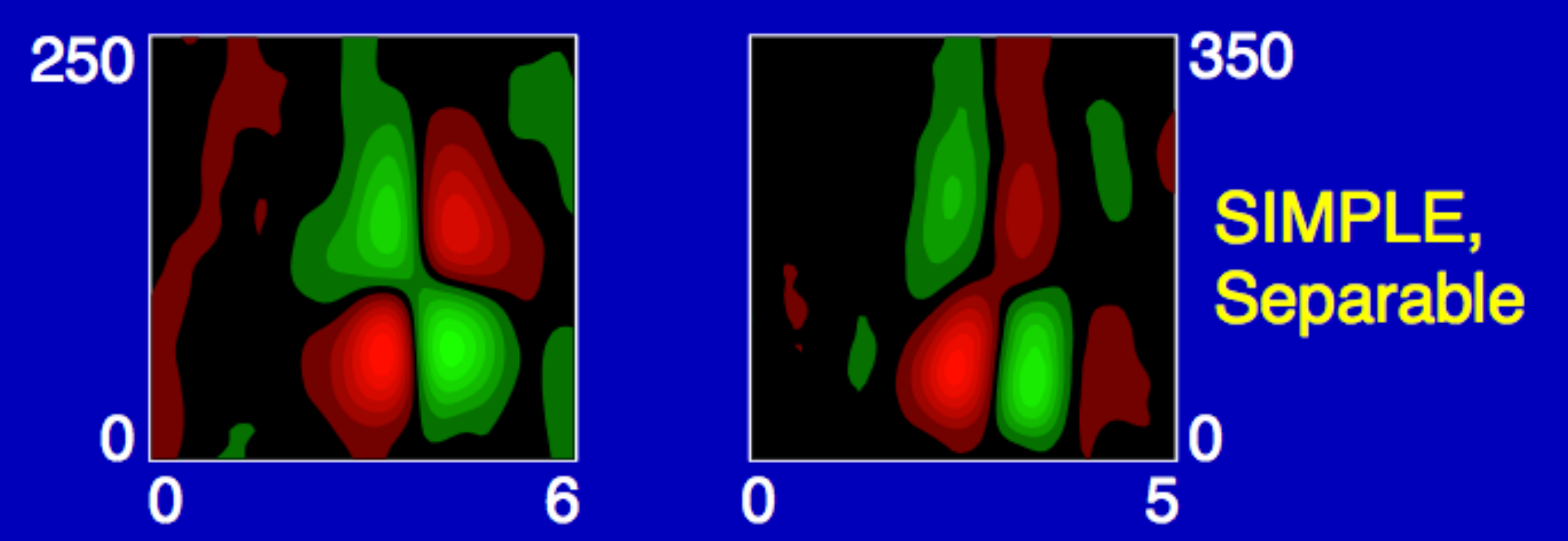} 
        & \includegraphics[width=0.48\textwidth]{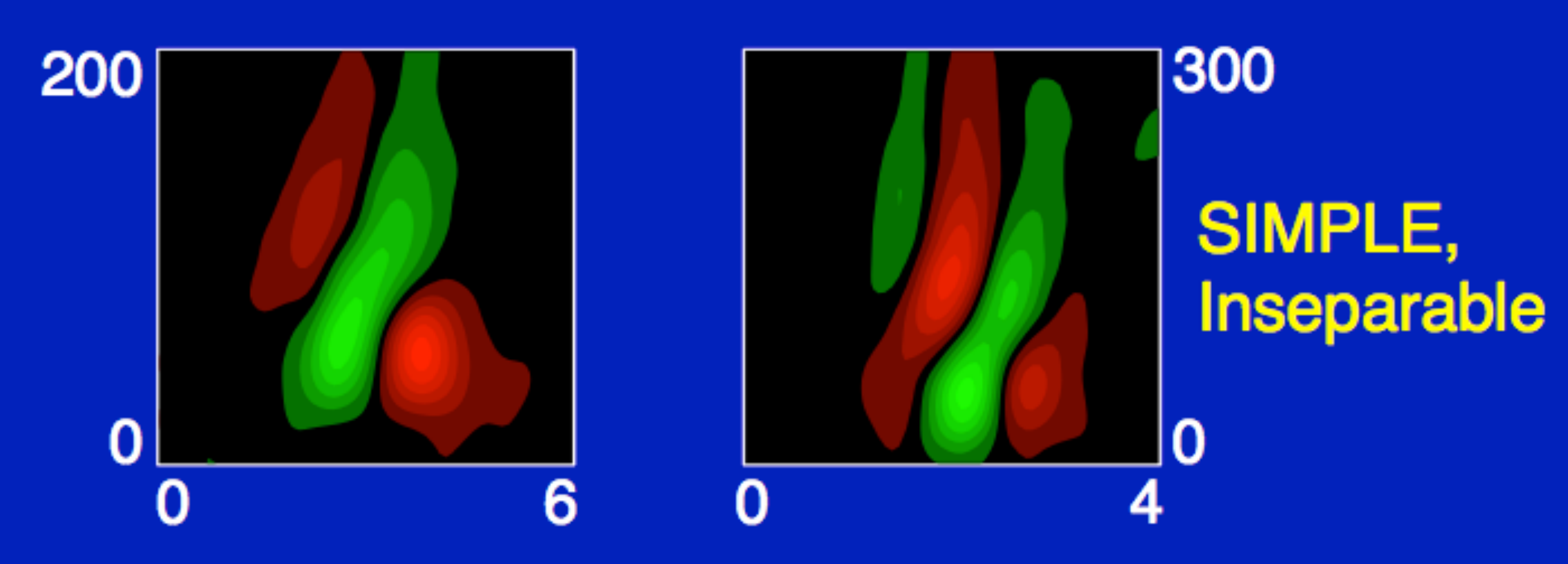}
    \end{tabular}
  \end{center}
  \begin{center}
    \begin{tabular}{ccccc}
      \hspace{-2mm} {\footnotesize $h_{xt}(x, t;\; s, \tau)$}
      & {\footnotesize $-h_{xxt}(x, t;\; s, \tau)$} 
      & \hspace{15mm}
     & \hspace{-1mm} {\footnotesize $h_{xx}(x, t;\; s, \tau, v)$}
      & {\footnotesize $-h_{xxx}(x, t;\; s, \tau, v)$} \\
      \hspace{-2mm} \includegraphics[width=0.15\textwidth]{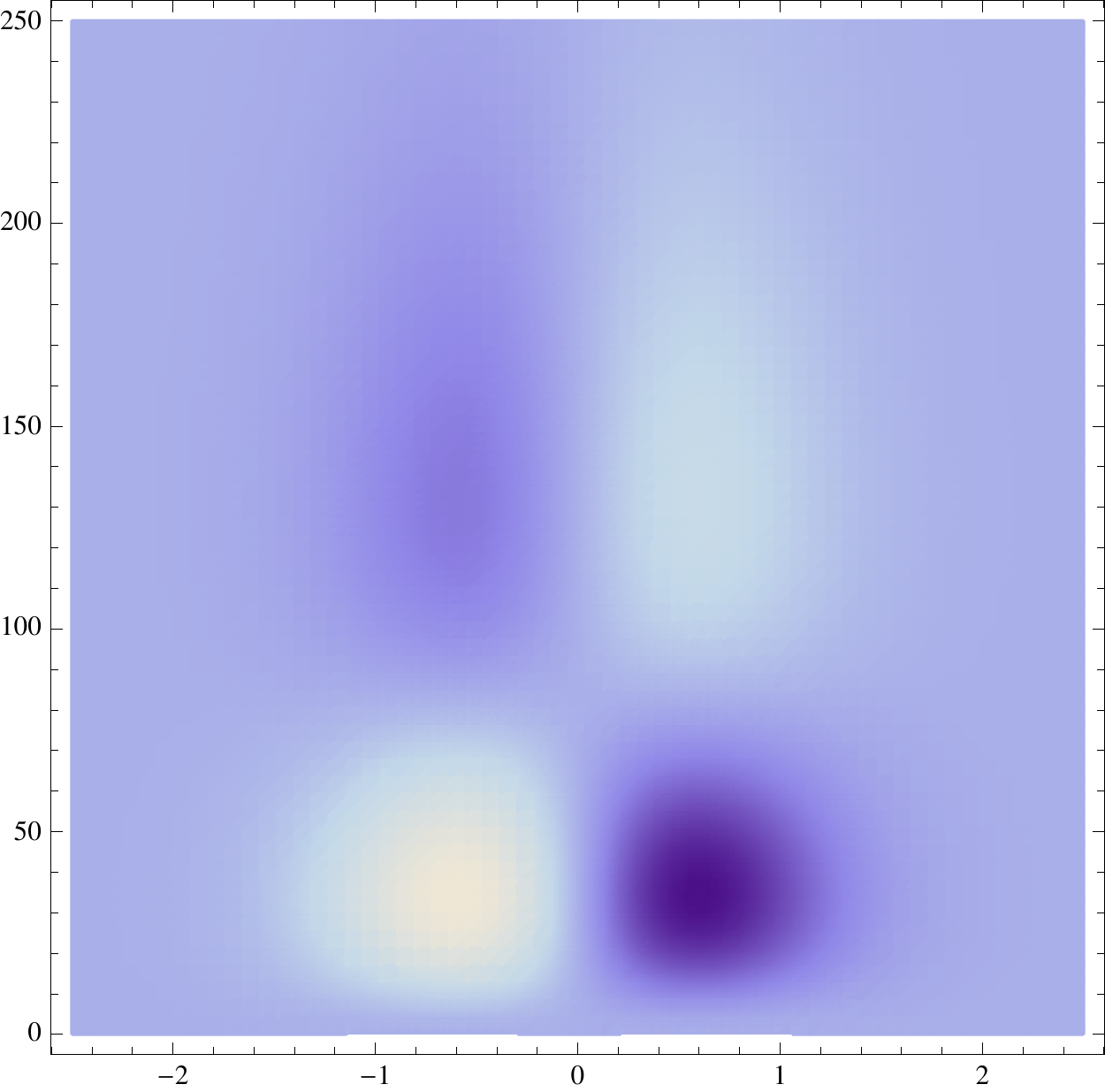}
      & \includegraphics[width=0.15\textwidth]{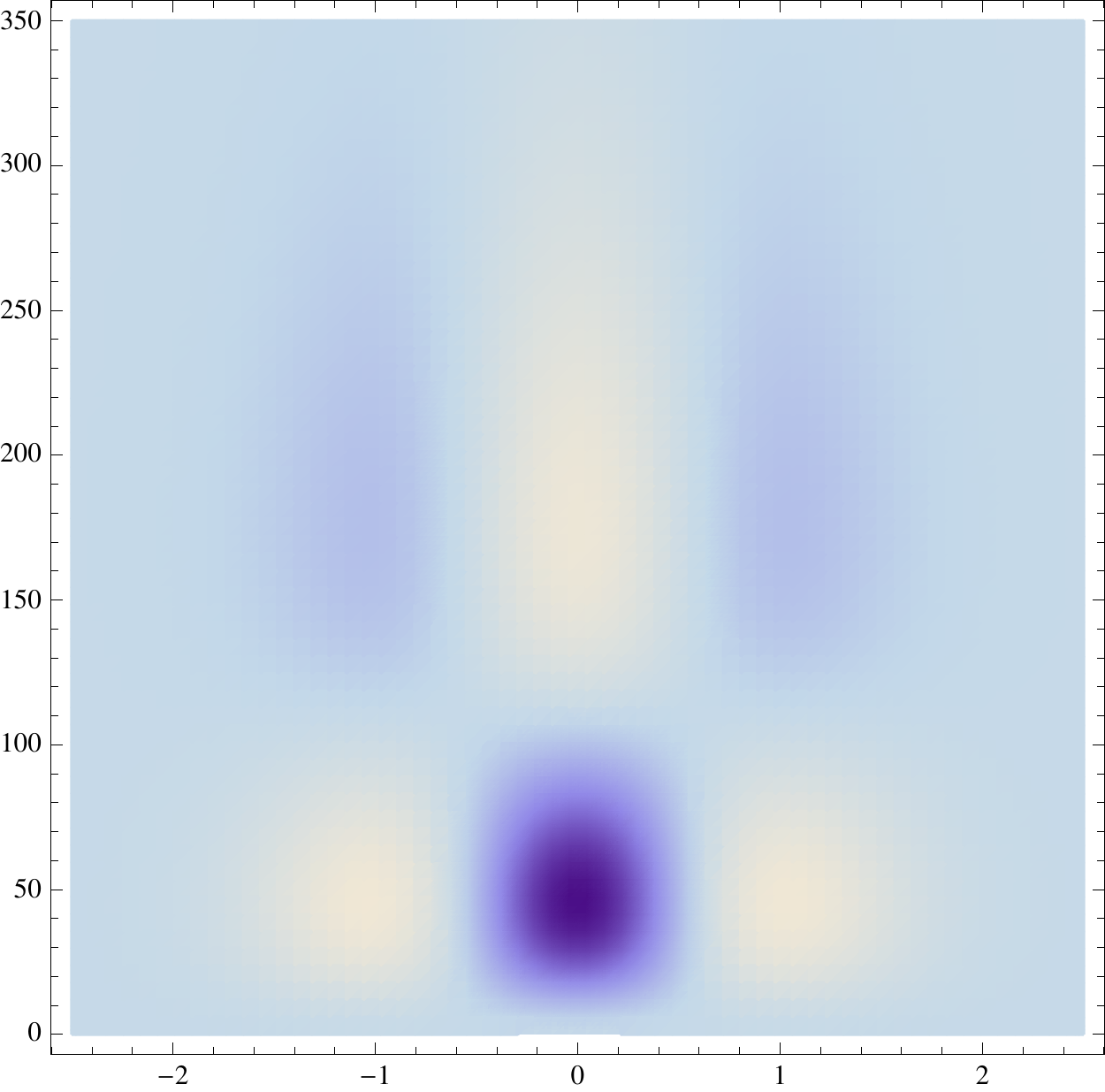} 
      & \hspace{15mm}
      & \hspace{-1mm} \includegraphics[width=0.15\textwidth]{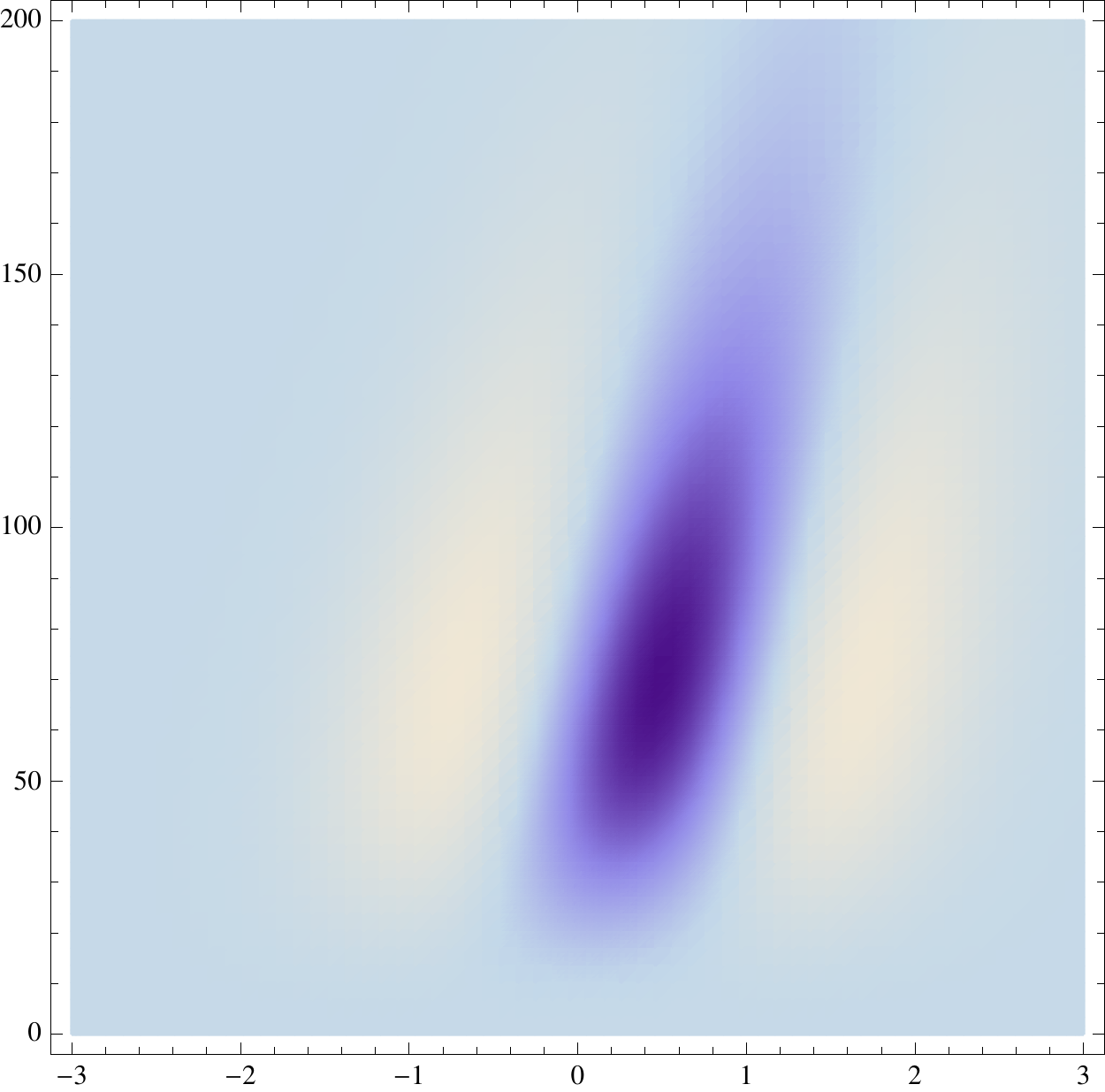}
      & \includegraphics[width=0.15\textwidth]{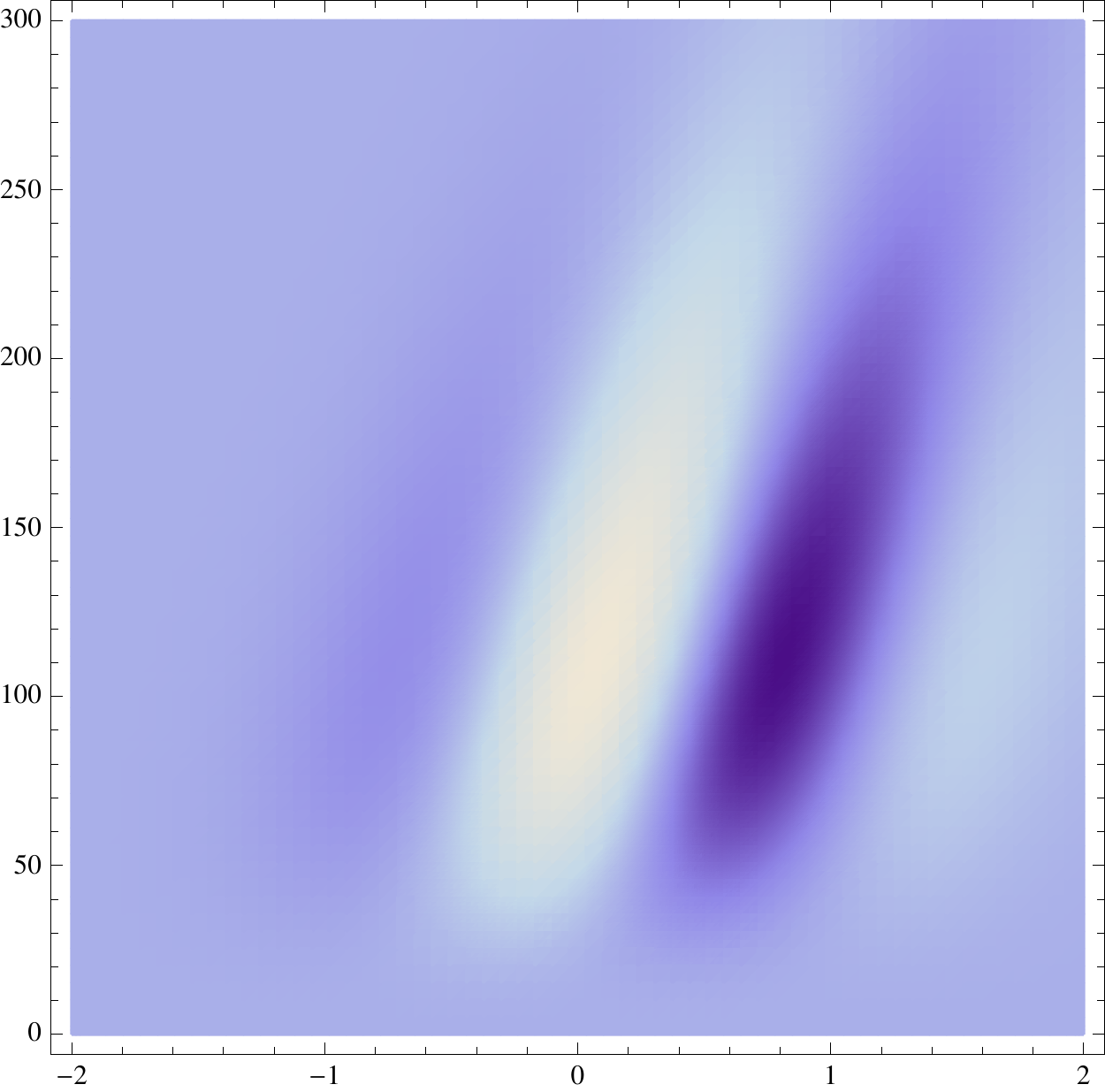}
    \end{tabular}
  \end{center}

  \caption{Computational modelling of simple cells in V1 as reported
    by DeAngelis et al.\ \cite{DeAngOhzFre95-TINS} using
    spatio-temporal receptive fields of the form
     $T(x, t;\; s, \tau, v)
           = \partial_{x^{\alpha}} \partial_{t^{\beta}} g(x - v t;\; s) \, h(t;\; \tau)$
    according to equation~(\protect\ref{eq-spat-temp-RF-model}) with the temporal smoothing function $h(t;\; \tau)$
    modelled as a cascade of first-order integrators/truncated exponential
    kernels of the form (\protect\ref{eq-comp-trunc-exp-cascade})
    and using a logarithmic distribution of the intermediate scale
    levels of the form (\protect\ref{eq-distr-tau-values}).
    (left column) Separable receptive fields corresponding to mixed
    derivatives of first- or second-order derivatives over space with
    first-order derivatives over time.
    (right column) Inseparable velocity-adapted receptive fields
    corresponding to second- or third-order derivatives over space.
    Parameter values: 
    (a) $h_{xt}$: $\sigma_x = 0.6$~degrees, $\sigma_t = 60$~ms.
    (b) $h_{xxt}$: $\sigma_x = 0.6$~degrees, $\sigma_t = 80$~ms.
    (c) $h_{xx}$: $\sigma_x = 0.7$~degrees, $\sigma_t = 50$~ms, $v = 0.007$~degrees/ms.
    (d) $h_{xxx}$: $\sigma_x = 0.5$~degrees, $\sigma_t = 80$~ms, $v = 0.004$~degrees/ms.}
   \label{fig-biol-model-simple-cells-rec-filters-over-time}
\end{figure}

When implementing this temporal scale-space concept, a set of
intermediate scale levels $\tau_k$ has to be distributed between some minimum
and maximum scale levels $\tau_{min} = \tau_1$ and $\tau_{max} = \tau_K$.
Assuming that a total number of $K$ scale levels is to be used, it is
natural to distribute the temporal scale levels according to a geometric series,
corresponding to a uniform distribution in units of
effective temporal scale $\tau_{eff} = \log \tau$.
Using such a logarithmic distribution of the temporal scale levels,
the different levels in the temporal scale-space representation at
increasing temporal scales will serve as a logarithmic memory of the
past, with qualitative similarity to the mapping of the
past onto a logarithmic time axis in the scale-time model by Koenderink \cite{Koe88-BC}.
If we have the freedom of choosing $\tau_{min}$ freely, a natural way
of parameterizing these temporal scale levels using a distribution
parameter $c > 1$ 
\begin{equation}
  \label{eq-distr-tau-values}
  \tau_k = c^{2(k-K)} \tau_{max} \quad\quad (1 \leq k \leq K)
\end{equation}
which by equation~(\ref{eq-mean-var-trunc-exp-filters}) implies that time
constants of the individual first-order integrators will be given by
\begin{align}
  \begin{split}
     \label{eq-mu1-log-distr}
     \mu_1 & = c^{1-K} \sqrt{\tau_{max}}
  \end{split}\\
  \begin{split}
     \label{eq-muk-log-distr}
     \mu_k & = \sqrt{\tau_k - \tau_{k-1}} = c^{k-K-1} \sqrt{c^2-1} \sqrt{\tau_{max}} \quad\quad (2 \leq k \leq K)
  \end{split}
\end{align}
If the temporal signal is on the other hand given at some minimum
temporal scale level $\tau_{min}$, 
we can instead determine $c = \left( \frac{\tau_{max}}{\tau_{min}} \right)^{\frac{1}{2(K-1)}}$
 in (\ref{eq-distr-tau-values})
such that $\tau_1 = \tau_{min}$ 
and add $K - 1$ temporal scales with $\mu_k$ according to (\ref{eq-muk-log-distr}).
Alternatively, if one chooses a uniform distribution of the
intermediate temporal scale levels 
\begin{equation}
  \label{eq-distr-tau-values-uni}
  \tau_k = \frac{k}{K} \, \tau_{max}
\end{equation}
then the time constants are given by
$\mu_k = \sqrt{\frac{\tau_{max}}{K}}$.

Figure~\ref{fig-trunc-exp-kernels-1D} shows graphs of such kernels 
that correspond to the same value of the composed variance,
using either a uniform distribution or a logarithmic distribution of
the intermediate scale levels.
Generally, these kernels are highly asymmetric for small values
of $K$, whereas they become gradually more symmetric as $K$ increases.
The degree of continuity at the origin and the smoothness of
transition phenomena increase with $K$ such that
coupling of $K \geq 2$ kernels in cascade implies a
$C^{K-2}$-continuity of the temporal scale-space kernel.
Specifically, the kernels based on a logarithmic
distribution of the intermediate scale levels allow for faster
temporal dynamics compared to the kernels based on a uniform
distribution of the intermediate scale levels.

Figure~\ref{fig-biol-model-simple-cells-rec-filters-over-time} shows
the result of modelling the spatio-temporal receptive fields of simple
cells in V1 in this way, using the general idealized model of
spatio-temporal receptive fields in equation~(\protect\ref{eq-spat-temp-RF-model})
in combination with a temporal smoothing kernel obtained by coupling a
set of first-order integrators/truncated exponential kernels in cascade.
This result complements the general theoretical model for visual
receptive fields in \cite{Lin13-BICY} with explicit modelling results
and a theory for
choosing and parameterizing the intermediate discrete temporal scale levels.

\section{Temporal dynamics of the time-causal kernels}
\label{app-temp-dyn}

For the time-causal filters obtained by coupling truncated exponential
kernels in cascade, there will be an inevitable temporal delay 
depending on the time constants $\mu_k$ of the individual filters.
A straightforward way of estimating this delay is by using the
additive property of mean values under convolution
  $m_K = \sum_{k=1}^K \mu_k$ 
according to (\ref{eq-mean-var-trunc-exp-filters}).
In the special case when all the time constants are equal $\mu_k =
\sqrt{\tau/K}$, this measure is given by
\begin{equation}
  \label{eq-delta-recfilt-uni}
  m_{uni} = \sqrt{K \tau} 
\end{equation}
showing that the temporal delay increases if the temporal smoothing
operation is divided into a larger number of smaller individual smoothing steps.

In the special case when the intermediate temporal scale levels are
instead distributed logarithmically according to (\ref{eq-distr-tau-values}), 
with the individual time constants given by
(\ref{eq-mu1-log-distr}) and (\ref{eq-muk-log-distr}),
this measure for the temporal delay is given by
\begin{align}
  \begin{split}
  \label{eq-delta-recfilt-log}
  m_{log} 
   & = \frac{c^{-K} \left(c^2-\left(\sqrt{c^2-1}+1\right) c+\sqrt{c^2-1} \, c^K\right)}{c-1} \, \sqrt{\tau }
  \end{split}
\end{align}
with the limit value
  $m_{log-limit} = \lim_{K \rightarrow \infty} m_{log} 
   = \frac{\sqrt{c^2-1}}{c-1} \sqrt{\tau}$
when the number of filters tends to infinity.

By comparing equations (\ref{eq-delta-recfilt-uni}) and
(\ref{eq-delta-recfilt-log}) including the limit value of the latter,
we can specifically note that with increasing number of intermediate
temporal scale levels, a logarithmic distribution of the intermediate
scales implies shorter temporal delays than a uniform
distribution of the intermediate scales.

Table~\ref{tab-tmean-uni-log-compare} shows numerical values
of these measures for different values of $K$ and $c$.
As can be seen, the logarithmic
distribution of the intermediate scales allows for significantly
faster temporal dynamics than a uniform distribution.

\begin{table*}[!hbt]
  \begin{center}
   \footnotesize
  \begin{tabular}{ccccc}
  \hline
   \multicolumn{5}{c}{Temporal mean values of time-causal kernels} \\
  \hline
    $K$ & $m_{uni}$  & $m_{log}$  ($c = \sqrt{2}$) & $m_{log}$  ($c = 2^{3/4}$) & $m_{log}$ ($c = 2$) \\
  \hline
    2 & 1.414 & 1.414 & 1.399 & 1.366 \\
    3 & 1.732 & 1.707 & 1.636 & 1.549 \\
    4 & 2.000 & 1.914 & 1.777 & 1.641 \\
    5 & 2.236 & 2.061 & 1.860 & 1.686 \\
    6 & 2.449 & 2.164 & 1.910 & 1.709 \\
    7 & 2.646 & 2.237 & 1.940 & 1.721 \\
    8 & 2.828 & 2.289 & 1.957 & 1.732 \\
%  16 & 4.000 & 2.406 & 1.732 \\
%  64 & 8.000 & 2.414 & 1.732 \\
  \hline
  \end{tabular}
\end{center}
\caption{Numerical values of the {\em temporal delay
    in terms of the temporal mean\/} $m = \sum_{k=1}^K \mu_k$
    in units of $\sigma = \sqrt{\tau}$ for time-causal kernels obtained by coupling $K$ truncated
    exponential kernels in cascade in the cases of a uniform
    distribution of the intermediate temporal scale levels $\tau_k = k \tau/K$ or a
    logarithmic distribution $\tau_k = c^{2(k-K)} \tau$ with $c > 1$.}
  \label{tab-tmean-uni-log-compare}

\begin{center}
   \footnotesize
  \begin{tabular}{ccccc}
  \hline
   \multicolumn{5}{c}{Temporal delays from the maxima of time-causal kernels} \\
  \hline
    $K$ & $t_{max,uni}$  & $t_{max,log}$ ($c = \sqrt{2}$) & $t_{max,log}$ ($c = 2^{3/4}$) & $t_{max,log}$ ($c = 2$) \\
  \hline
    2 & 0.707 & 0.707 & 0.688 & 0.640 \\
    3 & 1.154 & 1.122 & 1.027 & 0.909 \\
    4 & 1.500 & 1.385 & 1.199 & 1.014 \\
    5 & 1.789 & 1.556 & 1.289 & 1.060 \\
    6 & 2.041 & 1.669 & 1.340 & 1.083 \\
    7 & 2.268 & 1.745 & 1.370 & 1.095 \\
    8 & 2.475 & 1.797 & 1.388 & 1.100 \\
  \hline
  \end{tabular}
\end{center}
\caption{Numerical values for the {\em temporal delay of the local maximum\/} in
  units of $\sigma = \sqrt{\tau}$
    for time-causal kernels obtained by coupling $K$ truncated
    exponential kernels in cascade in the cases of a uniform
    distribution of the intermediate temporal scale levels $\tau_k = k \tau/K$ or a
    logarithmic distribution $\tau_k = c^{2(k-K)} \tau$ with $c > 1$.}
  \label{tab-tmax-uni-log-compare}
\vspace{-4mm}
\end{table*}

\paragraph{Additional temporal characteristics.}

Because of the asymmetric tails of the time-causal temporal smoothing
kernels, temporal delay estimation by the mean value may however lead
to substantial overestimates compared to {\em e.g.\/} the position of the local maximum.
To provide more precise characteristics in the case of a uniform
distribution of the intermediate temporal scale levels, for which
a compact closed form expression is available for the composed kernel
\begin{equation}
 \label{eq-composed-all-mu-equal}
  h_{composed}(t;\; \mu, K) = \frac{t^{K-1} \, e^{-t/\mu}}{\mu^K \, \Gamma(K)}
\end{equation}
let us differentiate this function
    $\partial_t \left( h_{composed}(t;\; \mu, K) \right)
    = \frac{e^{-\frac{t}{\mu }} ((K-1) \mu -t) \left(\frac{t}{\mu
        }\right)^{K+1}}{t^3 \, \Gamma(K)}$
and solve for the positions of the local maximum
\begin{align}
  \begin{split}
     \label{eq-tmax-recfilt-uni}
     t_{max,uni} & = (K-1) \, \mu 
                = \frac{(K-1) }{\sqrt{K}} \sqrt{\tau}.
  \end{split}
\end{align}
Table~\ref{tab-tmax-uni-log-compare} shows numerical values for the
position of the local maximum for both types of time-causal kernels.
As can be seen from the data, the temporal response properties are
significantly faster for a logarithmic distribution of the
intermediate scale levels compared to a uniform distribution
and the difference increases rapidly with $K$.
These temporal delay estimates are also significantly shorter than the
temporal mean values, in particular for the logarithmic distribution.

If we consider a temporal event that occurs as a step function over
time ({\em e.g.\/} a new object appearing in the field of view) and 
if the time of this event is estimated from the local maximum over
time in the first-order temporal derivative response, 
then the temporal variation in the response over time will be given by
the shape of the temporal
smoothing kernel. The local maximum over time will occur at a time
delay equal to the time at which the temporal kernel has its maximum
over time. Thus, the position of the maximum over time of the
temporal smoothing kernel is highly relevant for quantifying the
temporal response dynamics. 

\section{Computational implementation}
\label{sec-comp-impl}

The computational model for spatio-temporal receptive fields presented here
is based on spatio-temporal image data that are assumed to be
continuous over time.
When implementing this model on sampled video data,
the continuous theory must be transferred to discrete space and
discrete time.

In this section we describe how the temporal and spatio-temporal
receptive fields can be implemented in terms of corresponding discrete
scale-space kernels that possess scale-space properties over discrete
spatio-temporal domains.

\subsection{Discrete temporal scale-space kernels based on
  recursive filters}
\label{app-disc-temp-smooth}

Given video data that has been sampled by some temporal
frame rate $r$, the temporal
scale $\sigma_t$ in the continuous model in units of seconds is first 
transformed to a variance $\tau$ relative to a unit time sampling
\begin{equation}
  \label{eq-transf-tau-sampl}
  \tau = r^2 \, \sigma_t^2
\end{equation}
where $r$ may typically be either 25 Hz or 50 Hz.
Then, a discrete set of intermediate temporal scale levels $\tau_k$ is defined by
(\ref{eq-distr-tau-values})
or (\ref{eq-distr-tau-values-uni})
with the difference between successive scale levels according to 
  $\Delta \tau_k = \tau_k - \tau_{k-1}$ (with $\tau_0 = 0$).

For implementing the temporal smoothing operation between two such
adjacent scale levels (with the lower level in each pair of adjacent
scales referred to as $f_{in}$ and
the upper level as $f_{out}$), we make use of a {\em first-order recursive filter\/}
\begin{equation}
  \label{eq-norm-update}
  f_{out}(t) - f_{out}(t-1)
  = \frac{1}{1 + \mu_k} \,
    (f_{in}(t) - f_{out}(t-1))
\end{equation}
with generating function 
\begin{equation}
  \htransf_{geom}(z) = \frac{1}{1 - \mu_k \, (z - 1)}
\end{equation}
which is a time-causal kernel and satisfies discrete
scale-space properties of guaranteeing that the number of local extrema
or zero-crossings in the signal will not increase with increasing scale
(Lindeberg \cite{Lin90-PAMI}; Lindeberg and Fagerstr{\"o}m \cite{LF96-ECCV}).
Each such filter has temporal mean value $m_k = \mu_k$ and temporal variance
$\Delta \tau_k = \mu_k^2 + \mu_k$, and we compute $\mu_k$ from 
$\Delta \tau_k$ according to
\begin{equation}
  \mu_k = \frac{\sqrt{1 + 4 \Delta \tau_k}-1}{2}.
\end{equation}
By the additive property of variances under convolution,
the discrete variances of
the discrete temporal scale-space kernels will perfectly match those
of the continuous model, whereas the mean values and the temporal
delays may differ somewhat. 
If the temporal scale $\tau_k$ is large relative to the
temporal sampling density, the discrete model should be a good approximation in this respect.

By the time-recursive formulation of this temporal scale-space
concept, the computations can be performed based on a
compact temporal buffer over time, which contains the temporal
scale-space representations at temporal scales $\tau_k$ and with
no need for storing any additional temporal buffer of what
has occurred in the past to perform the corresponding temporal operations.

\subsection{Discrete implementation of spatial Gaussian smoothing}
\label{app-disc-gauss-smooth}

To implement the spatial Gaussian operation on discrete sampled data, we do first
transform a spatial scale parameter $\sigma_x$ in units of {\em e.g.\/} degrees 
of visual angle to a spatial variance $s$ relative to 
a unit sampling density according to
\begin{equation}
  \label{eq-transf-tau-sampl-s}
  s = p^2 \sigma_x^2
\end{equation}
where $p$ is the number of pixels per spatial unit {\em e.g.\/} in
terms of degrees of visual angle at the image center. Then, we
convolve the image data with the separable two-dimensional {\em discrete analogue of the
Gaussian kernel\/} (Lindeberg \cite{Lin90-PAMI}) 
\begin{equation}
  T(n_1, n_2;\; s) = e^{-2s} I_{n_1}(s)  \, I_{n_2}(s)
\end{equation}
where $I_n$ denotes the modified Bessel functions of integer order
and which corresponds to the solution of the semi-discrete diffusion
equation
\begin{equation}
  \partial_s L(n_1, n_2;\; s) =
  \frac{1}{2} (\nabla_{\times}^2 L) (n_1, n_2;\; s) 
\end{equation}
where $\nabla_{\times}^2$ denotes the five-point discrete Laplacian
operator defined by
$(\nabla_{\times}^2 f)(n_1, n_2) = 
f(n_1-1, n_2) + f(n_1+1, n_2) +f(n_1, n_2-1) + f(n_1, n_2+1)- 4 f(n_1, n_2)$.
These kernels constitute the natural way to define
a scale-space concept for discrete signals corresponding to the
Gaussian scale-space over a symmetric domain in the sense of
guaranteeing non-enhancement of local extrema, while
also ensuring a semi-group property
  $T(\cdot, \cdot;\; s_1) * T(\cdot, \cdot;\; s_2) = T(\cdot, \cdot;\; s_1 + s_2)$
over the discrete domain which implies that representations at coarser
scales can be computed from representations at finer scales using a
cascade property.

This operation can be implemented either by explicit spatial
convolution with spatially truncated kernels 
$\sum_{n_1=-N}^{N} \sum_{n_2=-N}^{N}  T(n_1, n_2;\; s) > 1 - \varepsilon$ 
for small $\varepsilon$ of the order $10^{-8}$ to $10^{-6}$ with
mirroring at the image boundaries (adiabatic boundary conditions)
or using the closed-form expression of the Fourier transform
$\varphi(\theta_1, \theta_2) 
= \sum_{n_1=-\infty}^{\infty} \sum_{n_1=-\infty}^{\infty} T(n_1,
n_2;\; s) \, e^{-i (n_1 \theta_1 + n_2 \theta_2)}
= e^{-2 t(\sin^2(\frac{\theta_1}{2}) +\sin^2(\frac{\theta_2}{2}))}$.

% Based on the (exact) relation 
% $\sum_{n=-\infty}^{\infty} T(n;\, s) = 1$,
% we truncate the infinite discrete kernel at the tails
% %\begin{equation}
%   %\label{eq-trunc-disc-gauss-tails}
%   $\sum_{n=-N}^{N} T(n;\; s) > 1 - \varepsilon$
% %\end{equation}
% for some small value of $\varepsilon$ of the order $10^{-8}$ to $10^{-6}$.
% For points where some part of the kernel stretches outside the
% domain of available data, we mirror the data at the boundaries,
% equivalent to solving the diffusion equation with
% adiabatic boundary conditions --- {\em i.e.\/} no heat transfer across
% the image boundaries.

\subsection{Discrete implementation of spatio-temporal receptive fields}

For separable spatio-temporal receptive fields, we implement the
spatio-temporal smoothing operation by separable combination of the
spatial and temporal 
scale-space concepts in sections~\ref{app-disc-temp-smooth} and 
\ref{app-disc-gauss-smooth}.
From this representation, spatio-temporal derivative
approximations are then computed from {\em difference operators\/} 
\begin{align}
  \begin{split}
     \delta_t & = (-1, +1) \quad\quad\quad
     \delta_{tt} = (1, -2, 1)
  \end{split}\\
 \begin{split}
     \delta_{x} & = (-\frac{1}{2}, 0, +\frac{1}{2}) \quad\quad
     \delta_{xx} = (1, -2, 1)
  \end{split}
\end{align}
expressed over the appropriate
dimension.
From the general theory in (Lindeberg \cite{Lin93-JMIV}) it follows that
the scale-space properties for the original 
zero-order signal will be transferred to such derivative
approximations, thereby implying theoretically well-founded
implementation of discrete derivative approximations.

For non-separable spatio-temporal receptive fields corresponding
to a non-zero image velocity $v = (v_1, v_2)^T$,
we implement the spatio-temporal smoothing operation by first warping 
the video data 
$(x_1', x_2')^T = (x_1 - v_1 t, x_2 - v_2 t)^T$
using spline interpolation. Then, we apply separable spatio-temporal
smoothing in the transformed domain and unwarp the result back to the
original domain.
Over a continuous domain, such an operation is equivalent to
convolution with corresponding velocity-adapted spatio-temporal
receptive fields, while being significantly faster in a discrete
implementation than corresponding explicit convolution with
non-separable receptive fields over three dimensions.

In addition to a transfer of the scale-space properties from the
continuous model to the discrete implementation, 
all the components in this discretization, the discrete Gaussian
kernel, the time-recursive filters and the discrete derivative
approximations, can be seen as mathematical approximations of the
corresponding continuous counterparts.
Thereby, the behaviour of the discrete implementation
will approach the 
corresponding continuous model.

\section{Summary and discussion}

We have presented an improved computational model for spatio-temporal
receptive fields based on time-causal and time-recursive
spatio-temporal scale-space representation defined from a set of
first-order integrators or truncated exponential filters
coupled in cascade over the temporal domain in combination with a Gaussian
scale-space concept over the spatial domain.
This model can be efficiently implemented in terms of recursive
filters and we have shown how the continuous model can be
transferred to a discrete implementation while retaining discrete
scale-space properties. Specifically, we have analysed how remaining
design parameters within the theory, in terms of the number of
first-order integrators coupled in cascade and a distribution
parameter of a logarithmic distribution, affect the temporal response
dynamics in terms of temporal delays.

Compared to other spatial and temporal scale-space representations
based on continuous scale parameters, a conceptual difference with the
temporal scale-space representation underlying the proposed
spatio-temporal receptive fields, is that the temporal scale levels
have to be discrete.
Thereby, we sacrifice scale invariance as resulting from Gaussian
scale-space concepts based on causality or non-enhancement of local
extrema (Koenderink \cite{Koe84}; Lindeberg \cite{Lin10-JMIV}) or 
used as a scale-space axiom in certain axiomatic scale-space formulations
(Iijima \cite{Iij62}; Florack et al.\ \cite{FloRom92-IVC}; 
Pauwels et al.\ \cite{PauFidMooGoo95-PAMI}; 
Weickert et al.\ \cite{WeiIshImi96-ScSp}; Duits et al.\
\cite{DuiFloGraRom04-JMIV}; Fagerstr{\"o}m \cite{Fag07-ScSp});
see also Koenderink and van Doorn \cite{KoeDoo90-BC},
Florack et al.\ \cite{FloRomKoeVie92-ECCV} and 
Tschirsich and Kuijper \cite{TscKui15-JMIV} for other scale-space
approaches closely related to this work.
For a vision system intended to operate in real time using no other explicit
storage of visual data from the past than a compact time-recursive
buffer of spatio-temporal scale-space at different
temporal scales, the loss of a continuous temporal scale parameter may however be less of a practical constraint,
since one would anyway have to discretize the temporal scale levels in
advance to be able to register the image data to perform any computations at all.

In the special case when all the time constants of the first-order
integrators are equal, the resulting temporal smoothing kernels 
in the continuous model (\ref{eq-composed-all-mu-equal})
correspond to Laguerre functions,
which have been previously used for modelling the temporal response
properties of neurons in the visual system (den Brinker and Roufs \cite{BriRou92-BC})
and for computing spatio-temporal image features in
computer vision (Rivero-Moreno and Bres \cite{RivBre04-ImAnalRec};
Berg et al.\ \cite{BerReyRid14-SensMEMSElOptSyst}). 
Regarding the corresponding discrete model with all time constants
equal, the corresponding discrete temporal smoothing kernels approach
Poisson kernels when the number of temporal smoothing steps increases
while keeping the variance of the composed kernel fixed
(Lindeberg and Fagerstr{\"o}m \cite{LF96-ECCV}).
Such Poisson kernels have also been used for modelling
biological vision (Fourtes and Hodgkin \cite{FouHod64-JPhys}).
Compared to the special case with all time constants equal,
a logarithmic distribution of the intermediate temporal scale levels (\ref{eq-distr-tau-values})
does on the other hand allow for larger flexibility in the trade-off between temporal smoothing and
temporal response characteristics, specifically enabling faster
temporal responses (shorter temporal delays) and higher computational efficiency when computing
multiple temporal or spatio-temporal receptive field responses involving coarser temporal scales.

%{\footnotesize
%\bibliographystyle{splncs}
%\bibliography{bib/defs,bib/tlmac}}

\end{document}